\documentclass[journal]{IEEEtran}

\usepackage{multirow}

\ifCLASSINFOpdf
  \usepackage[pdftex]{graphicx}
\else
\fi
\usepackage{tabularx,booktabs}

\usepackage{enumitem}
\usepackage{amsmath}
\usepackage{amssymb}
\usepackage[svgnames]{xcolor}
\usepackage{algorithm}
\usepackage{algpseudocode}

\usepackage{hyperref}

\usepackage{subcaption}
\makeatletter
\def\@makefnmark{\hbox to 2pt{$^{\@thefnmark}$\hss}}
\makeatother

\begin{document}
%
\title{Questions Beyond Pixels: \\Integrating Commonsense Knowledge in Visual Question Generation for Remote Sensing}
%
%
%

\author{Siran Li{*},~\IEEEmembership{IEEE Student Member},
        Li Mi{*},~\IEEEmembership{IEEE Student Member}, \\
        Javiera Castillo-Navarro, 
        Devis Tuia,~\IEEEmembership{IEEE Fellow}
\thanks{S. Li, L. Mi, J. Castillo-Navarro and D. Tuia are with Ecole Polytechnique Fédérale de Lausanne (EPFL). S. Li is also with the University of Tübingen. J. Castillo-Navarro is also with Conservatoire National des Arts et Métiers.}
\thanks{* indicates that authors contributed equally. Corresponding author: Li Mi, E-mail: li.mi@epfl.ch.}
}

%
%

\markboth{Journal of \LaTeX\ Class Files,~Vol.~14, No.~8, August~2015}%
{Shell \MakeLowercase{\textit{et al.}}: Questions beyond Pixels: Integrating Commonsense Knowledge in Visual Question Generation for Remote Sensing}
%



\maketitle

\begin{abstract}

With the rapid development of remote sensing image archives, asking questions about images has become an effective way of gathering specific information or performing semantic image retrieval. However, current automatically generated questions tend to be simplistic and template-based, which hinders the deployment of question answering or visual dialogue systems for real-world applications. To enrich and diversify the questions with both image content and commonsense knowledge, we propose a Knowledge-aware Remote Sensing Visual Question Generation model (KRSVQG). The proposed model incorporates related knowledge triplets from external knowledge sources to broaden the question content, while employing image captioning as an intermediary representation to ground questions to the corresponding images. Moreover, KRSVQG utilizes a vision-language pre-training and fine-tuning strategy, enabling the model's adaptation to low data regimes. To evaluate the proposed KRSVQG model, we construct two knowledge-aware remote sensing visual question generation datasets: the NWPU-300 dataset and the TextRS-300 dataset. Evaluations, including metrics and human assessment, demonstrate that KRSVQG outperforms existing methods and leads to rich questions, grounded in both image and domain knowledge. As a key practice in vision-language research, knowledge-aware visual question generation advances the understanding of image content beyond pixels, facilitating the development of knowledge-enriched vision-language systems with vision-grounded human commonsense.
\end{abstract}


%
\IEEEpeerreviewmaketitle

\section{Introduction}
%
%
%
%

\IEEEPARstart{R}{emote} sensing images contain a wealth of valuable information about the Earth's surface, but extracting useful information from the large image archives remains challenging. An effective strategy is the use of precise questions~\cite{mostafazadeh2016generating}. For example, query (\textit{``What are on the water and parked beside the bridge?"}), should direct models to efficiently identify images with boats. Visual Question Generation (VQG) for Remote Sensing Images (RSVQG) aims to generate such specific questions about remote sensing images, which plays a pivotal role in actively interacting with remote sensing data and provides possibilities to build a powerful Visual Question Answering (VQA) system or a Visual Dialog system~\cite{lobry2020rsvqa}.

\begin{figure}[t]
    \centering
    \includegraphics[width=7.5 cm]{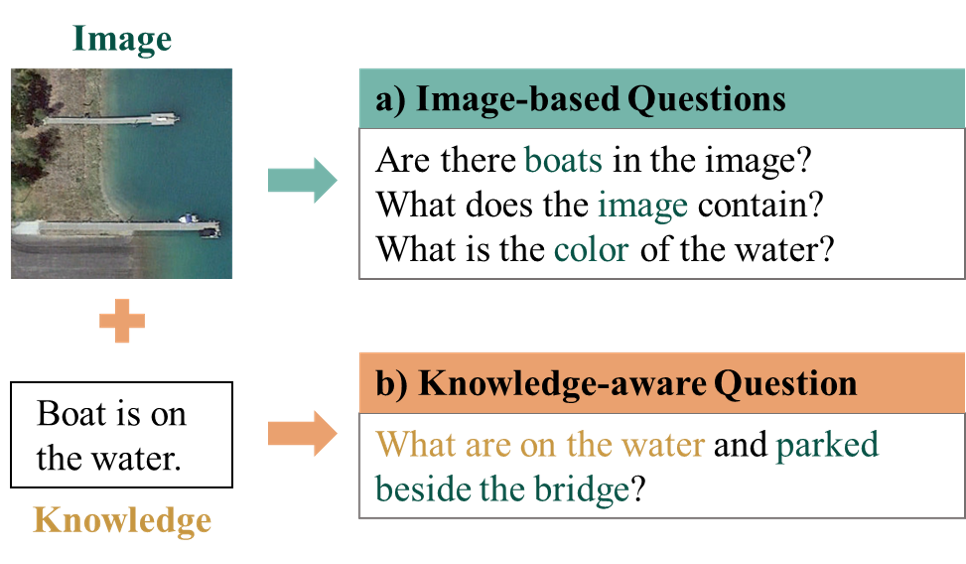}
    \caption{An example of \textbf{image-based questions} and a \textbf{knowledge-aware question}. In the questions, text related to image description is highlighted in green, while text related to external knowledge is in orange.}
    \label{fig:sample}
\end{figure}

The quality and diversity of the questions generated are crucial to the success of such systems. Unfortunately, 
conventional image-based question generation systems tend to generate questions that are redundant, template-based, and that mostly focus on the presence of objects, rather than incorporating broader context or real-world information~\cite{bashmal2023visual}. Examples of such systems are shown in the green box of Fig.~\ref{fig:sample}, where generated questions are about objects (\textit{Are there boats in the image?}) or very generic (\textit{What does the image contain?}). 
Those questions are valid but generic -- they can be asked about any image. With those questions, it is difficult to meet the requirements of extracting information of interest and exploring commonsense beyond the raw image content (\textit{e.g.}, about the function of objects). 
\textcolor{Black}{In this paper, we introduce Knowledge-aware Visual Question Generation for Remote Sensing images, which integrates external knowledge into VQG tasks. This integration facilitates the generation of questions requiring complex reasoning and informative insights~\cite{gao2022cric}.} This integration not only enhances the contextual understanding of visual content, but also fosters specificity in questioning~\cite{xie2022knowledge, uehara2023k}. For example, the knowledge-aware question in the orange box of Fig.~\ref{fig:sample} enhances specificity and relevance with respect to the image. It extracts location information (\textit{parked beside the bridge}) directly from the visual content and incorporates commonsense knowledge, such as (\textit{boats are typically found on water.}).

However, integrating commonsense knowledge into remote sensing images understanding still faces challenges~\cite{mi2022knowledge}. Unlike the generalized commonsense knowledge bases used in natural language contexts, there are few comprehensive knowledge bases tailored specifically for remote sensing~\cite{li2021robust}. Therefore, effective strategies must be developed to extract domain-specific knowledge from generalized sources and adapt general vision-language models for the remote sensing domain.

To generate knowledge-enriched questions for remote sensing images, we propose a Knowledge-aware RSVQG method. The proposed method incorporates external knowledge to enrich the generated questions, while leveraging descriptive captions as intermediary representations, and therefore enhancing the image grounding of the generated questions. In addition, to accommodate limited labeled data, the proposed model adopts a vision-language pre-training and fine-tuning strategy: 1) a vision pre-training stage to adapt the vision module to the remote sensing field, 2) a language pre-training step to facilitate knowledge integration, and 3) a fine-tuning stage on target datasets with limited data annotations to generate knowledge-aware questions. To evaluate KRSVQG, we create two datasets with 600 manually annotated samples in total (NWPU-300 and TextRS-300). The datasets are created by integrating external knowledge from the ConceptNet~\cite{speer2017conceptnet} database into questions, resulting in questions characterized by comprehensive and diverse content.

Our main contributions can be summarized as follows:
\begin{itemize}
\item We propose KRSVQG, a method for knowledge-aware visual question generation specifically tailored to remote sensing images. Our approach generates question-answer pairs that incorporate both knowledge information and visual content. Additionally, a training strategy including pre-training and fine-tuning is used to adapt the model with limited data annotations in remote sensing.
\item We create and make available two knowledge-aware remote sensing VQG datasets, NWPU-300 and TextRS-300. They are based on existing captioning datasets, each containing 300 manually annotated samples. Each sample in the dataset is composed of an image, a caption, a knowledge triplet, and a question-answer pair, offering valuable resources for advancing research in this domain. 
The data and code are available at the link: \url{https://github.com/Siran-Li/KRSVQG}.
\end{itemize}

A preliminary study of this work was published by the authors in a conference paper~\cite{li2024knowledge}. We extend it by: 1) providing a detailed description and analysis of the proposed datasets (Section~\ref{datasets}); 2) presenting our proposed training strategy (Section~\ref{methodology}.C); 3) adding two state-of-the-art question generation baselines on the proposed datasets (Section~\ref{results}.A); 4) providing an in-depth methodological analysis, including training strategy analysis, parameter analysis, and human evaluation (Section~\ref{results}.B and D). 
The rest of this paper is organized as follows: Section~\ref{related_work} reviews related work on VQG and knowledge-aware vision-language research. Section~\ref{datasets} details the sources and construction process of the proposed datasets. Section~\ref{methodology} describes the proposed KRSVQG method, including the model architecture and the training strategy. Section~\ref{setup} and Section~\ref{results} present the experimental setup and results, respectively. Finally, Section~\ref{conclusion} concludes the paper.

\section{Related Work \label{related_work}}

\subsection{Visual Question Generation Methods}

VQG is a particular case of Question Generation (QG)~\cite{mitkov2003computer}, integrating information of visual content for generating questions~\cite{mostafazadeh2016generating}. Initially, QG methods relied heavily on templates to generate questions from text~\cite{mitkov2003computer}. Thanks to the adoption of pre-trained language models (\textit{e.g.}, BERT~\cite{devlin2019bert} and T5~\cite{raffel2020exploring}), and large-scale datasets \textit{(e.g.}, QG-Bench~\cite{ushio2022generative}), the generation of questions becomes context-aware and follows a logical sequence~\cite{zhang2019addressing}.

Early VQG models generated questions only based on the input images. Mora \textit{et al.}~\cite{mora2016towards} used a CNN-LSTM model to directly generate question-answer pairs given an image. Li \textit{et al.}~\cite{li2018visual} incorporated answers as input and treated VQG and VQA as a dual task to formulate questions and answers. Krishna \textit{et al.}~\cite{krishna2019information} developed an IM-VQG model that extracts targeted information from images for question generation by reconstructing answers and question categories. While these models could generate image-based questions, they primarily produced questions based on pre-defined templates and often encountered challenges in achieving diverse question content. 

Beyond the image content, the incorporation of commonsense knowledge empowers VQG models to tap into external information, facilitating the generation of informative and knowledge-aware questions. Xie \textit{et al.}~\cite{xie2022knowledge} introduced non-visual knowledge related to the answer into the VQG task. Uehara \textit{et al.}~\cite{uehara2023k} utilized a multi-modal transformer to encode both visual information and target knowledge, generating questions based on the fused features. Mi \textit{et al.}~\cite{mi2024convqg} proposed two modality-specific contrastive objectives to ground the generated question to both image content and external knowledge. However, those models proposed for knowledge-aware question generation cannot be directly used for the remote sensing domain due to the lack of annotation data.
 
Despite the extensive research in VQG for natural images, the exploration of VQG in the remote sensing domain is still in its early stages. Bashmal \textit{et al.}~\cite{bashmal2023visual} proposed a paragraph-based VQG approach that employed a vision transformer (ViT) to extract visual features from images and generate various questions using the GPT-2 model. While the generated questions cover several objects detected in remote sensing images, the content of the questions cannot be controlled, resulting in some general questions, yes/no, or numeric questions. In this paper, inspired by the literature on knowledge-aware question generation in natural images~\cite{uehara2023k, mi2024convqg}, we integrate external commonsense knowledge bases to enhance the quality and specificity of questions generated for remote sensing images. In addition, we also design a specific training strategy to adapt knowledge-aware question generation methods to the remote sensing domain with less annotation data.

\subsection{Visual Question Generation Datasets}
As a dual task with VQA, VQG can be evaluated on VQA datasets~\cite{mora2016towards, li2018visual, bashmal2023visual}. The VQA2.0~\cite{goyal2017making} and Visual7W~\cite{zhu2016visual7w} datasets are widely employed for VQG tasks for training models to generate questions related to the images. Moving from image-based questions to knowledge-aware questions, the K-VQG dataset~\cite{uehara2023k}, a large-scale human-annotated dataset, emerges as a significant resource. Each sample in the dataset is annotated with a target knowledge triplet and a corresponding knowledge-aware question. 

Compared with the various VQA datasets for natural images, there are only a few VQA datasets specifically constructed for remote sensing images. \textcolor{Black}{The RSVQA-LR, RSVQA-HR datasets~\cite{lobry2020rsvqa}, and RSIVQA dataset~\cite{zheng2021mutual} are answer-centric datasets. Questions in those datasets are created using templates, resulting in fixed question patterns that the model can use to create shortcuts and eventually solve the task without even accessing the image~\cite{chappuis2023curse}.} TextRS-VQA~\cite{bashmal2023visual}, while not restricted to fixed question types, still focuses on image-based questions with simple and generic content.

Acknowledging the constraints of existing datasets for the RSVQG task, we opted to create our evaluation datasets based on the existing remote sensing image captioning datasets and to leverage the information contained in the captions. The NWPU caption dataset~\cite{cheng2022nwpu} is a large-scale remote sensing dataset with diverse captions describing image content, and the TextRS dataset~\cite{abdullah2020textrs} contains captions with generic descriptions focusing on the main objects in the image. Based on these two captioning datasets, we developed our knowledge-aware datasets through the integration of external knowledge into the question, resulting in questions characterized by comprehensive and diverse content (see Section~\ref{datasets}).

\subsection{Vision-Language Pretraining}

Early methods in remote sensing vision-language research are often based on CNN and RNN architectures, employing a CNN to capture image features and an RNN to handle the language aspects~\cite{lobry2020rsvqa, krishna2019information, mora2016towards}. These methods laid the foundation for integrating visual and textual information, forming a bridge between image understanding and language comprehension.

In recent developments, vision-language pre-training has attracted attention by providing robust image and text representation~\cite{li2019visualbert, radford2021learning, li2022blip}. 
For example, the BLIP model~\cite{li2022blip} adopts a multimodal architecture, enabling the model to jointly consider textual and visual inputs. Such models have demonstrated effectiveness in various vision-language tasks~\cite{zhu2023chatgpt}. 

In our KRSVQG model, we employed BLIP as a base model, leveraging its vision-language understanding capabilities gained through pre-training on large-scale vision-language datasets. To adapt the model to limited annotations, we proposed a vision-language pre-training and fine-tuning strategy (Section~\ref{sec:training}). 
This approach ensures the generation of nuanced, knowledge-aware questions that improve the model's ability to represent and describe remote sensing images.

\section{Datasets \label{datasets}}

To evaluate the proposed KRSVQG method, we build two knowledge-aware remote sensing VQG datasets. The proposed datasets are based on samples from two existing remote sensing image captioning datasets: 300 images from the NWPU dataset~\cite{cheng2022nwpu} and 300 from the TextRS dataset~\cite{abdullah2020textrs}. Each data sample consists of the following components: (1) image $I$, (2) caption $C$, (3) knowledge sentence $S$, (4) question $Q$, and (5) answer $A$, where the images and captions are from the original remote sensing image captioning datasets. In this section, we introduce our data sources, the dataset construction process, and present the basic statistics of the proposed datasets.

\subsection{Data Sources}

\textbf{Image Source}. 
The NWPU caption dataset~\cite{cheng2022nwpu} features 31,500 images, each with 5 detailed captions describing the images captured from various sensors, with resolutions between $30-0.2$m. The TextRS dataset~\cite{abdullah2020textrs} contains 2,144 images, with 2-5 captions for each image, but these captions offer more general descriptions of the predominant objects. Since the captions of the NWPU dataset are longer and with detailed information about the image, we randomly selected one caption for each image. For the TextRS dataset, the caption is shorter, so we combine all the sentences into a single one to get more information on the image content. The 300 images are selected randomly across various scene classes (\textit{e.g.} agricultural, beach, harbor, etc.) for diversity and comprehensiveness. Specifically, the images of the NWPU-300 dataset for the training and validation datasets are drawn from the original training and validation sets of the NWPU caption dataset separately to prevent data leakage in the fine-tuning process (See Section~\ref{sec:training}).

\textbf{Knowledge sources}.
Our knowledge source is ConceptNet~\cite{speer2017conceptnet}, a graph designed to represent commonsense knowledge and relationships between concepts. It contains over 1.7 million entities and 3.4 million edges as relationships. A typical triplet in ConceptNet is represented as \textit{$<$head, relationship, tail$>$}, such as \textit{$<$mobile houses, at location, street$>$}.
We chose 14 specific types of relations from ConceptNet, following previous works on knowledge-aware VQG task~\cite{xie2022knowledge, uehara2023k}, leading to a set of relations $R$, including \{\textit{has a, used for, capable of, at location, has subevent, has prerequisite, has property, causes, created by, defined as, desires, made of, not desires, receives action}\}. \textcolor{Black}{By introducing external concepts or relations from ConceptNet, the proposed datasets introduce human commonsense (e.g., trees provide shade from the sun, hills are used for climbing, etc.), which is often implicit or absent in the RS visual content.}

\begin{figure*}[t]
    \centering
    \includegraphics[width=14cm]{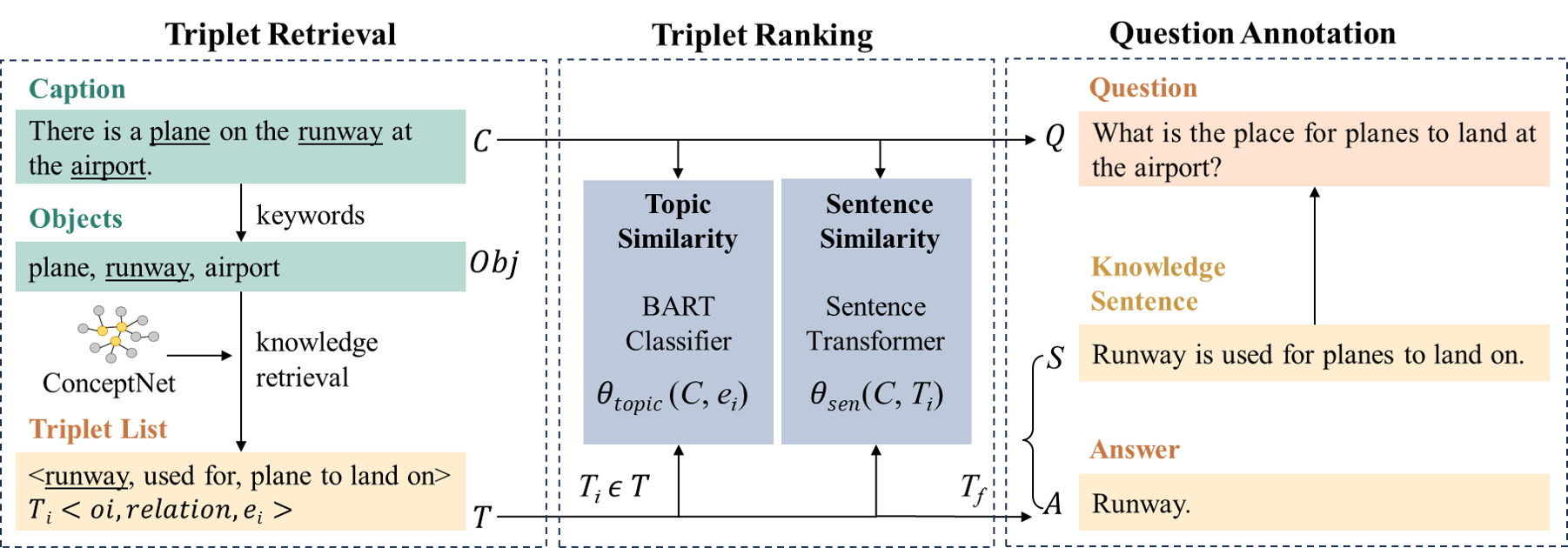}
    \caption{The annotation process for the NWPU-300 and TextRS-300 datasets involves three stages: First, Triplet Retrieval, where the knowledge triplets related to the original caption are retrieved from ConceptNet. Next, Triplet Ranking, where the retrieved triplets are ranked based on their similarity to the caption. Finally, Question Annotation, where we annotate a question based on the selected triplet.
    }
    \label{fig:construction}
\end{figure*}

\subsection{Dataset Construction}\label{data-construction}

Based on the aforementioned data sources, our annotation process, follows three main steps (Fig.~\ref{fig:construction}):

\textbf{Triplet Retrieval}. For an image $I$ with a caption $C$, we extract a list of $n$ nouns from $C$, shown as $Obj=\{o_1, o_2, ..., o_n\}$. Subsequently, based on the object list, we retrieve the one-step neighborhood of the objects and collect the retrieved knowledge triplet list from ConceptNet, labeled as $T$. Each triplet in the triplet list, $T_{i} \in T$, takes the form of $T_{i} \langle o_{i}, relation, e_{ij} \rangle$ or $T_{i}\langle e_{ij}, relation, o_{i}\rangle$, where $o_{i} \in Obj$ represents the selected object noun, and $e_{ij}$ represents the entity from the external knowledge source with $j$ indicating multiple possible connections, and \textit{relation} denotes one of the relations from $R$, connecting $o_{i}$ and $e_{ij}$. We also retain the noun chunks that contain the selected nouns along with any connected adjectives or numbers as potential answers $A$, since the additional context and descriptions surrounding the extracted nouns can provide specific information for the image. For example, from the phrase (\textit{A large ship floating in the water}), we extract the noun chunk \textit{A large ship} as Answer $A$, with \textit{ship} as the object $o_{i}$. 

\textbf{Triplet Ranking}. Depending on the number of object and the number of 1-step neighboring entities per object, as a list of triplets is created to keep only the most informative ones, we rank the triplets in two phases. Firstly, we assess the \textit{topic similarity} between the caption $C$ and the external entity $e_i$ using a pre-trained BART model~\cite{lewis2019bart}~\footnote{https://huggingface.co/facebook/bart-large-mnli}. The model is fine-tuned for natural language inference and used as a ready-made zero-shot sequence classifier~\cite{yin2019benchmarking}. This classifier works by predicting the similarity score between a given input text (in this case, the caption $C$) and a set of possible labels or classes (the external entities $e_{ij}$). The model generates a similarity score $\theta_{topic}(C, e_{ij})$, and if this score falls within the threshold range of [0.2, 0.8], we retain the corresponding triplet. This range filters out triplets that are either overly similar to the captions or have a significant semantic difference from the caption. This method effectively filters out irrelevant entities while identifying those that meaningfully relate to the caption.

Secondly, we measure the \textit{sentence similarity} between the caption $C$ and the triplet $T_i$ using a pre-trained Sentence Transformer~\cite{reimers2019sentence}~\footnote{https://huggingface.co/sentence-transformers/all-MiniLM-L6-v2}. The model is based on the transformer architecture, and fine-tuned to produce dense vector embeddings that capture the overall semantic meaning of sentences. By encoding both the caption $C$ and the triplet $T_i$ into vectors, the model allows us to compute a similarity score $\theta_{sen}(C, T_{i})$, reflecting how closely the triplet aligns with the caption semantically. To ensure relevance and avoid redundancy, we again set a threshold for this similarity score within the range of [0.2, 0.8]. We then sort the triplets in descending order by their similarity scores, prioritizing those most relevant to the caption. This ensures that the selected triplets offer supplementary and meaningful information.

\begin{table}[t]
\setlength{\tabcolsep}{5pt}
\centering
\caption{\textcolor{Black}{The template to convert the relations in the knowledge triplets from ConceptNet into knowledge sentences.}}

\resizebox{0.95 \columnwidth}{!}{
\begin{tabular}{llll}
\toprule
\textbf{Relation} & \textbf{Template}  & \textbf{Relation} & \textbf{Template}  \\
\midrule
UsedFor         & is used for         &  ReceivesAction  & receives action \\
HasA   & has a     &  Causes & causes \\
HasProperty   & has a property              &  CreatedBy & is created by \\
DefinedAs          & is defined as          & AtLocation & is at location of \\
HasSubEvent     & has      & MadeOf  & is made of \\
HasPrerequisite       & has prerequisite      & Desires & desires \\
NotDesires  & not desires & CapableOf      & is capable of \\

\bottomrule
\end{tabular}}

\label{tab:template}
\end{table}

\textbf{Annotation}. We manually select one triplet $T_{f}$ from the top-10 triplets from the list after ranking.
\textcolor{Black}{We then annotate a question ($Q$), a knowledge sentence ($S$) and an answer ($A$) for each sample. For the answer $A$, we used the noun chunk related to the object $o_{f}$ (\textit{i.e.}, the object along with any connected adjectives or numbers in the caption $C$). The knowledge sentence $S$ is automatically derived from the selected knowledge triplet $T_{f}$. Specifically, for each selected triplet $T_{f}$$<o_{f}, relation_{f}, e_{f}>$ or $T_{f}$$<e_{f}, relation_{f}, o_{f}>$, we first replace the object $o_{f}$ with the corresponding noun chunk. This replacement enhances the grounding of the knowledge sentence to the image content. After substitution, we convert the triplet into a natural language sentence. The triplet can be divided into three parts: head, relation, and tail. We keep the head and tail as the subject and object of the sentence, and rewrite the relation annotation according to a template as the predicate of the knowledge sentence. For example, the triplet $< boat, AtLocation, water >$ can be rewritten as “boat is at location of water”. The transformation templates are listed in Table~\ref{tab:template}.}
Based on $S$, $C$, and image $I$, we manually annotate a question $Q$ leading to the expected answer $A$. Note that after ranking and selection, the numbers of relation types in the two datasets are reduced because some relation types are filtered out in the process.

\begin{figure}[t]
\centering
  \centering
  \includegraphics[width=0.85\linewidth]{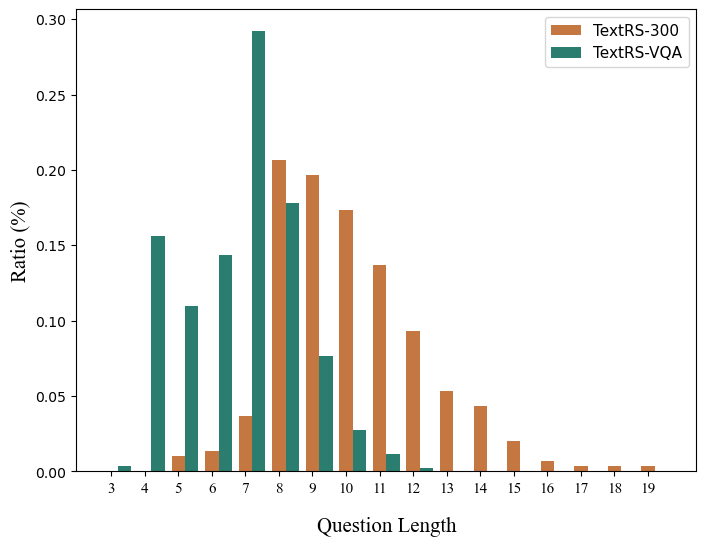}
\caption{The distribution of question lengths in the TextRS-300 dataset and TextRS-VQA dataset.}
\label{fig:data_statistics}
\end{figure}

\begin{table}[t]
    \caption{Comparison of vocabulary diversity of the annotated questions between TextRS-VQA and TextRS-300 datasets. Avg LenQ denotes the average length of the question, \#Nouns denotes the number of nouns, \#Verbs denotes the number of verbs, and \#Adjectives denotes the number of Adjectives.}
    \label{tab:diversity}
    \centering
    \begin{tabular}{ccccc}
    \toprule
        ~ & Avg LenQ & \#Nouns & \#Verbs & \#Adjectives \\ \midrule
        TextRS-VQA & 6.64 & 164 & 37 & 42 \\ 
        TextRS-300 & 10.05 & 226 & 106 & 51 \\ \bottomrule
    \end{tabular}
\end{table}

\begin{figure}[t]
    \centering
    \includegraphics[width=8.5cm]{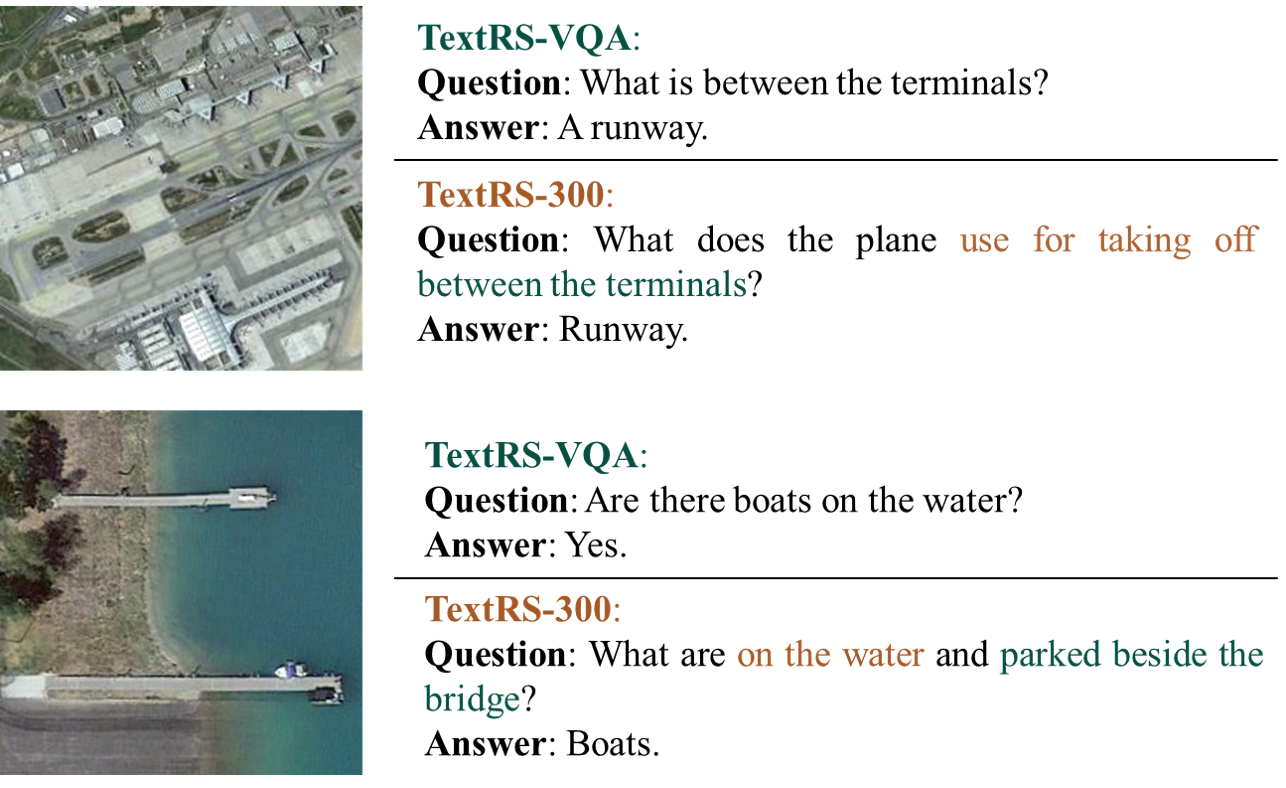}
    \caption{The examples from TextRS-VQA dataset and TextRS-300 dataset. In the questions, text from the caption description is highlighted in green, and text from external knowledge is highlighted in orange.}
    \label{fig:anno_example}
\end{figure}

\begin{table}[t]
    \caption{Comparison between the NWPU-300 and TextRS-300 datasets. \#I denotes the number of images; \#C/I: number of captions for each image; \#R: number of relations, Avg LenC: average length of the caption; Avg ObjC: average number of objects in the caption; Avg LenQ: average length of the question.}
    \label{tab:RS-anno}
    \centering
    \begin{tabular}{ccccccc}
    \toprule
        Dataset & \#I & \#C/I & \#R & Avg LenC & Avg ObjC & Avg LenQ \\ \midrule
        NWPU-300 & 300 & 5 & 8 & 13.34 & 3 & 10.09 \\ 
        TextRS-300 & 300 & 2-5 & 7 & 6.37 & 2 & 10.05 \\ \bottomrule
    \end{tabular}
\end{table}

\begin{table}[t]
\setlength{\tabcolsep}{5pt}
\centering
\caption{\textcolor{Black}{Comparison with RS-oriented knowledge base. \#Entities, \#Relations, and \#Triplets denote the counts of unique entities, relation types, and triplets, respectively.}}
\begin{tabular}{cccc}
\toprule
Knowledge base & \#Entities  & \#Relations & \#Triplets  \\
\midrule
RSKG~\cite{li2021robust} & 117         &  26  & 191 \\
NWPU-300 & 339 & 8 & 254 \\
TextRS-300 & 267 & 7 & 202 \\
\bottomrule
\end{tabular}
\label{tab:kg_counts}
\end{table}

\subsection{Dataset Statistics}

For the same set of 300 images, we compare the knowledge-aware questions in the proposed TextRS-300 dataset with the image-based questions in the TextRS-VQA dataset~\cite{bashmal2023visual}. To ensure a fair comparison, for each question in the TextRS-300 dataset, we randomly select a question corresponding to the same image from the TextRS-VQA dataset. Fig.~\ref{fig:data_statistics} summarises the question length distribution of the two datasets. In the TextRS-VQA dataset, most questions consist of 7 words. In contrast, the TextRS-300 dataset comprises longer questions, spanning from 8 to 14 words, with an average question length of 10. Additionally, we evaluate the vocabulary diversity in the annotated questions. As shown in Table~\ref{tab:diversity}, the TextIRS-300 dataset exhibits a higher vocabulary diversity than TextIRS-VQA, and a substantial increase in word types: 38\% more nouns, 186\% more verbs, and 21\% more adjectives. This observed distinction aligns with our expectations, as the TextRS-300 dataset incorporates more detailed and varied questions generated using knowledge and caption information.

We also present some annotation examples in Fig.~\ref{fig:anno_example}. In summary, questions and answers in the TextRS-VQA dataset tend to be more generic, including many yes/no questions, while the TextRS-300 dataset provides a narrower focus on specific objects and attributes present in the images. For instance, questions like \textit{What is between the terminals?} in the TextRS-VQA dataset lack specificity in target answers, encompassing possibilities like \textit{runway}, \textit{lawn}, or \textit{airplanes} for the given image. However, by integrating commonsense knowledge, the knowledge-aware question \textit{What does the plane use for taking off between the terminals?} limits the possible answer to objects that can be used for the plane to take off, \textit{e.g.}, \textit{runway}, \textit{apron}, or \textit{tarmac}. With the image content, the correct answer \textit{runway} can be inferred.

Going beyond the question, Table~\ref{tab:RS-anno} analyzes the statistics of both datasets. 
The NWPU-300 captions contain more nouns than the TextRS-300 dataset, and the average length is longer. Both datasets exhibit similar question lengths, with questions of around 10 words on average. It is noteworthy that the TextRS-300 dataset exhibits a significantly shorter average caption length, approximately half that of the NWPU-300 dataset.

\subsection{Comparison with RS-oriented knowledge base}
\textcolor{Black}{We compare the knowledge triplet annotations in the proposed NWPU-300 and TextRS-300 datasets with RSKG~\cite{li2021robust}, a widely used RS-oriented knowledge graph in Table ~\ref{tab:kg_counts}. 
RSKG includes 117 entities, 26 relations, and 191 triples, focusing primarily on structured representations of remote sensing concepts such as land cover types and spatial relationships. Compared to the RS-oriented knowledge graph, the proposed datasets include significantly more entities and triplets. However, they contain fewer relation types, as spatial relations (e.g., next to, near) and attribute-based relations (e.g., color, height) are not specified or explicitly presented as relations in the proposed datasets.}

\section{Methodology \label{methodology}}

\begin{figure*}[htp]
    \centering
    \includegraphics[width=18cm]{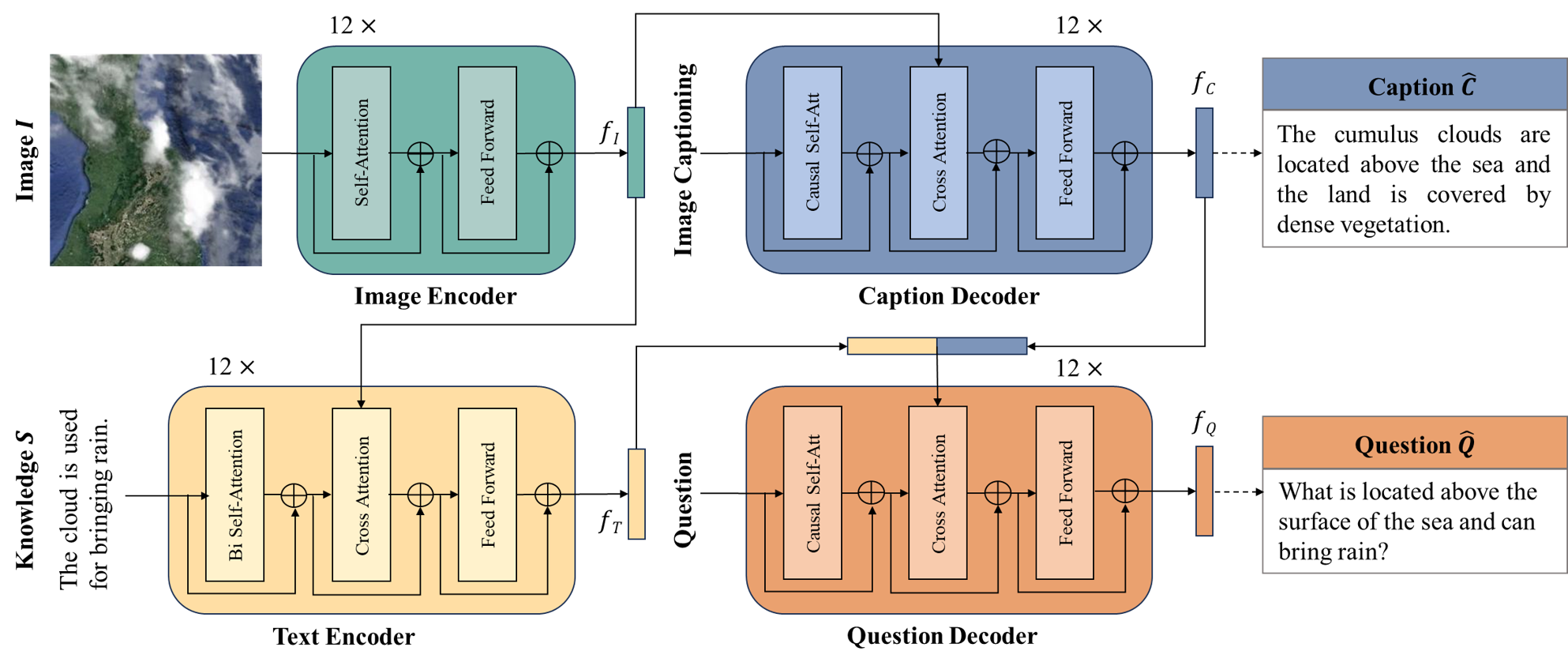}
    \caption{The KRSVQG model and its four components \textcolor{Black}{(See Section~\ref{sec:arch})}: (i) image encoder ($\mathbf{ViT}$); (ii) caption decoder ($\mathbf{BERT_{CapDec}}$); (iii) text encoder ($\mathbf{BERT_{TextEnc}}$); and (iv) question decoder ($\mathbf{BERT_{QueDec}}$). With an image ($I$) and a knowledge sentence ($S$) as inputs, the model generates a knowledge-aware question ($\widehat{Q}$) and caption ($\widehat{C}$) based on the image ($I$) and the knowledge sentence ($S$). \textcolor{Black}{KRSVQG is trained under the proposed training strategy (See Section~\ref{sec:training}): In vision pre-training (VPT), tunable parameters are confined to the vision module (upper row). During language pre-training (LPT), both vision and language modules are jointly trained. In the fine-tuning (FT) stage, the vision module initializes from VPT weights, while the language module initializes from LPT weights. Both modules are updated during FT.}  $\oplus$ means addition and normalization. \textcolor{Black}{Dotted arrows indicate the inference path.}}
    \label{fig:diagram}
\end{figure*}

To generate knowledge-enriched, informative questions, we propose a knowledge-aware visual question generation model for remote sensing images. \textcolor{Black}{Note that both images and knowledge sentences are necessary inputs in the task of knowledge-aware RSVQG.} Besides, to adapt the model to limited annotated data, we also introduce a specialized training strategy, involving two pre-training stages for vision and language separately, followed by a final fine-tuning stage on the target knowledge-aware RSVQG datasets.  

\subsection{Architecture}
\label{sec:arch}
Fig.~\ref{fig:diagram} shows the overall architecture of our proposed KRSVQG model. It is composed of two main modules: one pertaining to vision and another to language.

\subsubsection{Vision Module} 
KRSVQG is based on the BLIP structure~\cite{li2022blip} and comprises an image encoder to extract image feature representations. Besides image encoding, we also leverage caption generation as an intermediate stage. 

\textbf{Image Encoder}.
Given an input image $I$, an image encoder based on ViT~\cite{dosovitskiy2020image} encodes image features: $f_{I} = \mathbf{ViT}(I)$. Following BLIP, the classification token [CLS] of the ViT is used to represent the image feature.

\textbf{Caption Decoder}.
 The image feature vector is sent to a caption decoder, a variation of the BERT model~\cite{devlin2019bert}, to produce the caption feature: $f_{C} = \mathbf{BERT_{CapDec}}(f_{I})$. The caption decoder incorporates a causal self-attention layer~\cite{dong2019unified}, which attends only to the preceding tokens when predicting the next one. This ensures the model generates text in an autoregressive manner, maintaining logical consistency and grammatical correctness by restricting the focus to the current and past context. 
 In addition to the causal self-attention, the decoder also includes a cross-attention layer that injects the visual feature $f_{I}$ into the caption generation process. Together, these components enable the decoder to generate coherent and contextually relevant captions based on the input image features. The final caption $\widehat{C}$ is generated from the caption feature $f_{C}$.

\subsubsection{Language Module} 
We use the image-grounded text encoder in the BLIP model to process the input knowledge sentence and a final question decoder that fuses the image content with external knowledge to generate questions.

\textbf{Text Encoder}.
The text encoder is also based on the BERT model~\cite{devlin2019bert}. It employs bidirectional self-attention layers, which allow it to consider the entire context of the knowledge sentence \(S\) simultaneously. This bidirectional approach enables the encoder to build comprehensive representations by attending to both past and future tokens in the sentence. Moreover, the input knowledge sentence is fused with the image feature \(f_{I}\) through a cross-attention layer, which integrates visual and textual information: \(f_{T} = \mathbf{BERT_{TextEnc}}(S, f_{I})\). The resulting encoded feature \(f_{T}\) serves as a multimodal representation that encapsulates the information from the knowledge sentence and the associated visual context, thereby enabling the model to generate questions grounded in both sources.

\textbf{Question Decoder}.
The question decoder shares the same BERT-based structure as the caption decoder, designed to generate questions by combining the information from image content ($f_{C}$) and external knowledge ($f_{T}$). These fused features are injected into the decoder via a cross-attention layer: $f_{Q} = \mathbf{BERT_{QueDec}}(concat(f_{C}, f_{T}))$. The final question $\widehat{Q}$ is decoded from the feature representation $f_{Q}$, resulting in a context-aware question that reflects both the image and the knowledge sentence.

\subsection{Loss Functions}
In the KRSVQG model, the vision module generates captions from images, guided by a captioning loss. Meanwhile, the language module formulates questions by integrating visual and knowledge representations, monitored by a question generation loss. Both processes leverage cross-entropy loss to evaluate the dissimilarity between model predictions and target annotations.

The caption generation loss is:
\begin{equation}
\label{eq:cross-entropy}
\begin{aligned}
\mathbf{Loss}_{CG} = - \underset{n=1}{\sum^{|\widehat{C}|}}logP(\widehat{C}_{n}|\{\widehat{C}_{<n}\}),
\end{aligned}
\end{equation}
where $|\widehat{C}|$ is the number of tokens in $\widehat{C}$, $\widehat{C}_{n}$ is the token being generated at the $n$-th position, and $\widehat{C}_{<n}$ is the sequence of generated tokens up to the ($n-1$)-th step. Similarly, the question generation phase also utilizes a cross-entropy loss, formulated as:
\begin{equation}
\label{eq:qgcross-entropy}
\begin{aligned}
\mathbf{Loss}_{QG} = - \underset{n=1}{\sum^{|\widehat{Q}|}}logP(\widehat{Q}_{n}|\{\widehat{Q}_{<n}\}),
\end{aligned}
\end{equation}
where $|\widehat{Q}|$ is the number of tokens in $\widehat{Q}$, $\widehat{Q}_{n}$ is the token being generated at the $n$-th position, and $\widehat{Q}_{<n}$ is the sequence of generated tokens up to the ($n-1$)-th step.

\subsection{Training Strategy}\label{training-strategy}
\textcolor{Black}{For remote sensing images, the low data regimes are common as high-quality annotations require considerable efforts and expertise. Regarding this challenge, we exploit the capabilities of vision-language models pretrained on large-scale general domain datasets and tailor them to the RS domain. More specifically, we introduce a comprehensive vision-language pre-training and fine-tuning strategy for our KRSVQG model, including vision pre-training, language pre-training, and vision-language fine-tuning.}

\textbf{Vision Pre-training (VPT)}.
\label{sec:training}
\textcolor{Black}{VPT refers to pretraining the vision module on RS-specific content through remote sensing image captioning using the NWPU dataset~\cite{cheng2022nwpu}. This allows the model to adapt the visual feature representation to the remote sensing domain and benefit from the extensive annotations available in remote sensing image captioning datasets.
By refining the capabilities of both the image encoder and the caption decoder during VPT, we significantly improve the model's ability to learn domain-relevant visual representations from the outset.}

\textbf{Language Pre-training (LPT)}.
LPT aims to equip the KRSVQG model with the capability to generate questions based on the relevant knowledge sentence $S$. LPT uses knowledge-aware VQG as the pretraining task and the corresponding K-VQG dataset~\cite{uehara2023k}. The dataset is comprised of 16,098 samples, each including an image, a question-answer pair, and a knowledge triplet. Due to the absence of caption annotations in the K-VQG dataset, we leverage a pre-trained BLIP image captioning model~\cite{li2022blip}  to automatically generate image captions for K-VQG samples. We also simply concatenate the knowledge triplets to construct concise sentences, as described in section~\ref{data-construction}. \textcolor{Black}{Note that during LPT, we train all parameters of the KRSVQG model, including the image encoder, caption decoder, text encoder, and question decoder.}

\textbf{Vision-Language Fine-tuning (FT)}.
In the final stage, we fine-tune the entire KRSVQG model (Fig.~\ref{fig:diagram}). The FT is carried out separately on the proposed two knowledge-aware remote sensing VQG datasets. The FT has two phases: initially, as the vision module has been pre-trained on a remote sensing dataset and the language module on a natural image dataset, we freeze the weights of the vision module. This strategy aims to balance between the $\mathbf{Loss}_{CG}$ and $\mathbf{Loss}_{QG}$, avoiding large differences between the two loss values. Subsequently, we proceed with end-to-end fine-tuning of the entire KRSVQG model on the target task. 

\subsection{Inference}\label{inference}

\textcolor{Black}{During inference, the model follows a two-step sequential generation process. First, the image encoder extracts the image features $f_I$, which are passed to the caption decoder to get the hidden state feature $f_{C}$ and then generate the caption $\widehat{C}$, autoregressively. Second, $f_C$, and together with the knowledge embedding $f_T$ obtained from the text encoder, are concatenated and provided to the question encoder to get the hidden state feature $f_{Q}$, from which the final question $\widehat{Q}$ is generated autoregressively.}

\section{Experimental Setup \label{setup}}

\subsection{Experimental Setting}
In the experiments, we cover the following aspects:
\begin{itemize}[leftmargin=*]
\item {\textbf{Main Results}}. We report the results and compare KRSVQG with state-of-the-art methods.
\item {\textbf{Ablation Study}}. Ablation studies are performed for the proposed model architecture and training strategy. A further low-data regime analysis is performed to demonstrate the effectiveness of the proposed training strategy.
\item \textbf{Qualitative Analysis}. We analyze the quality and diversity of our generated questions. 
\item \textbf{Human Evaluation}. We perform a human evaluation to compare the quality of questions generated by the different approaches.
\end{itemize}

\subsection{Evaluation Metrics and Dataset Split}
Following the VQG evaluation in the literature~\cite{krishna2019information, bashmal2023visual}, we assess the performance of different QG models using the following metrics: 

\begin{itemize}[leftmargin=*]
\item BLEU (BiLingual Evaluation Understudy)~\cite{papineni2002bleu} is a metric that assesses n-gram matches between the generated text and the references. Variations of BLEU include BLEU-n (n=1, 2, 3, 4), based on the n-gram used in the comparison. 
\item METEOR (Metric for Evaluation of Translation with Explicit ORdering)~\cite{banerjee2005meteor} is an evaluation metric based on the harmonic mean of unigram precision and recall of the generated text according to the references, with recall weighted more than precision. 
\item ROUGE$_{L}$ (Recall-Oriented Understudy for Gisting Evaluation)~\cite{lin-2004-rouge} is a metric that compares the longest common subsequence between the model's output and the reference.
\item CIDEr (Consensus-based Image Description Evaluation)~\cite{vedantam2015cider} is a metric that treats each sentence as a document and calculates the Term Frequency-Inverse Document Frequency of the n-grams. It then computes the cosine similarity between the n-grams of the generated text and the references.
\end{itemize}

To facilitate model training and validation, we partition the NWPU-300 and TextRS-300 datasets into training and validation sets, maintaining a 4:1 ratio. Specifically, the training dataset comprises 240 samples, while the validation dataset contains 60 samples. 
We keep this train/validation split for the proposed datasets for all the experiments except for the low-data regime analysis. For VPT and LPT, we follow the dataset splits of the corresponding datasets in the literature.

\subsection{Implementation Details} 
The BLIP base model with ViT-B and CapFilt-L consists of 224M parameters, and the pre-trained weights are publicly accessible~\footnote{https://github.com/salesforce/BLIP}. \textcolor{Black}{For VPT and LPT, the model weights are initialized with BLIP pre-trained weights. Then, for the FT stage, the vision module is initialized with weights from the VPT pre-trained model, while the language module is initialized with weights from the LPT pre-trained model.} \textcolor{Black}{The image features ($f_{I}$) and the caption features ($f_{C}$) are represented in $\mathbb{R}^{B \times L_C \times dim}$, where $B$ is the batch size, $L_{C}$ is the maximum length of the caption and $dim$ is the feature dimension. Knowledge embeddings ($f_{T}$) are represented in $\mathbb{R}^{B \times L_T \times dim}$ and the final question features are represented in $\mathbb{R}^{B \times (L_C + L_T) \times dim}$. In practice, we set $L_{C} = L_{T} = 20$ and $dim = 768$.}

We utilize a single NVIDIA GeForce RTX 2080 Ti GPU for our experiments. The training epochs for VPT and LPT are set to 10, and for FT to 20, with a batch size of 1. The initial learning rate for the VPT and LPT is set to $2e^{-5}$, and for FT is set to $1e^{-6}$. We use the AdamW optimizer with a weight decay of 0.05. The images are resized to 384 $\times$ 384 during pre-training and fine-tuning.

\subsection{Baseline Methods}

For the knowledge-aware VQG task on remote sensing images, specific baseline models do not exist. Therefore, we adapt current QG and VQG models to knowledge-aware VQG for remote sensing, establishing the following four methods as baselines. \textcolor{Black}{To ensure a fair comparison, all the models are pre-trained on the K-VQG dataset~\cite{uehara2023k} (LPT) and fine-tuned for the proposed NWPU-300 and TextRS-300 datasets (FT) by default. Note that vision pre-training (VPT) is not applicable in the baseline methods since there is no caption generated in their models.} 
\begin{itemize}[leftmargin=*]
\item {IM-VQG}~\cite{krishna2019information}: The IM-VQG model is a VQG model that leverages variational auto-encoders to reconstruct the image and answer representations, enhancing the model's ability to generate questions that align with the provided answer while maximizing information from the image. For a fair comparison, we adapt the IM-VQG model for knowledge-aware visual question generation: We keep the image input but replace the original answer input with the knowledge sentence $S$ and guide the model to reconstruct the knowledge sentence to help generate knowledge-aware questions.
\item {LMQG}~\cite{ushio2022generative}: LMQG is a set of QG models fine-tuned on QG-Bench, a large-scale QG dataset. Those sequence-to-sequence models are designed to automatically generate questions from given paragraphs. In our experiments, we use a pre-trained T5-small model~\cite{raffel2020exploring} from the LMQG model set. For a fair comparison, the text input consists of the concatenated knowledge sentence and caption, as the model cannot process images directly.
\item {TextRS-VQG}~\cite{abdullah2020textrs}: The TextRS-VQG uses Swin-Transformer~\cite{liu2021swin} and GPT-2~\cite{radford2019language} as an encoder-decoder architecture. The model was proposed to generate general questions for remote sensing images. To ensure a fair comparison, we encode the knowledge sentence by a pre-trained Sentence Transformer~\cite{reimers2019sentence} and combine the encoded knowledge feature with the image feature extracted by ViT as the input of the GPT-2 model.
\item {ConVQG}~\cite{mi2024convqg}: ConVQG, also based on the BLIP architecture, leverages two modality-specific contrastive objectives to guide the content of the generated question by both image content and text constraints. It achieved state-of-the-art knowledge-aware VQG performance on several natural image benchmarks. We take the image and knowledge sentences as the visual and textual input, respectively.
\end{itemize}

\section{Experimental Results \label{results}}

\begin{table}[t!]
\caption{\label{tab:nwpu-caption} Vision pretraining results. We compare the image captioning performance of the vision module of KRSVQG (BLIP) with state-of-the-art image captioning models on the NWPU dataset. Best performance is in bold. }
\setlength\tabcolsep{3pt}
    \centering
    {\renewcommand{\arraystretch}{1.2}
    \begin{tabular}{cccccc}
    \toprule
        Model & BLEU-1 & BLEU-4 & METEOR & ROUGE$_{L}$ & CIDEr \\ 
        \midrule
        MLCA~\cite{cheng2022nwpu}& 74.50 & 47.80 & 33.70 & 60.10 & 1.26 \\ 
        RSDT~\cite{du2023plane} & 75.15 & 48.28 & 31.87 & 58.58 & 1.21 \\ 
        HMLP~\cite{wei2023remote} & 78.90 & 51.10 & 35.10 & 63.30 & 1.30 \\ 
        BLIP~\cite{li2022blip} & \textbf{90.90} & \textbf{75.92} & \textbf{48.28} & \textbf{82.60} & \textbf{2.11} \\ 
        \bottomrule
    \end{tabular}}
\end{table}

\begin{table}[t!]
\caption{\label{tab:finetune-kvqg} Language pre-training results. We compare the KRSVQG model with state-of-the-art QG and VQG models on the K-VQG dataset. The best performance is in bold, and "-" means that the result is not provided in the original paper.}
\setlength\tabcolsep{3pt}
    \centering
    {\renewcommand{\arraystretch}{1.2}
    \begin{tabular}{cccccc}
    \toprule
        Model & BLEU-1 & BLEU-4 & METEOR & ROUGE$_{L}$ & CIDEr \\ 
        \midrule
        
        IM-VQG~\cite{krishna2019information} & 45.68 & 12.33 & 16.45 & 43.17 & 0.37 \\ 
        LMQG~\cite{ushio2022generative} & 46.10 & 16.82 & 21.01 & 42.17 & 0.92 \\
        KVQG~\cite{uehara2023k} & - & 18.84 & \textbf{22.79} & - & 1.31 \\
        ConVQG~\cite{mi2024convqg} & 51.25 & 20.01 & 22.66 & \textbf{46.34} & \textbf{1.53} \\ 
        \textbf{KRSVQG} & \textbf{52.19} & \textbf{20.24} & 22.49 & 46.06 & 1.34 \\ 
        \bottomrule
    \end{tabular}}
\end{table}

\begin{table*}[t!]
    \centering
    {\renewcommand{\arraystretch}{1.2}
    \begin{tabular}{c|cccccccc}
    \toprule
        Dataset & Model & BLEU 1 & BLEU 2 & BLEU 3 & BLEU 4 & METEOR & ROUGE$_{L}$ & CIDEr \\
        \midrule
        & IM-VQG~\cite{krishna2019information} & 32.46 & 15.91 & 7.93 & 0.00 & 8.96 & 31.95 & 0.17 \\ 
        & TextRS-VQG~\cite{bashmal2023visual} & 29.84 & 17.09 & 11.25 & 7.67 & 14.08 & 32.39 & 0.45 \\ 
        NWPU-300 & LMQG~\cite{ushio2022generative} & 28.10 & 17.45 & 10.98 & 6.54 & 16.34 & 26.33 & 0.85 \\ 
        & ConVQG~\cite{mi2024convqg} & 34.13 & 22.59 & 16.86 & 13.16 & 15.69 & 37.96 & 1.00 \\ 
        & \textbf{KRSVQG} & \textbf{41.87} & \textbf{28.32} & \textbf{20.73} & \textbf{14.78} & \textbf{18.70} & \textbf{38.48} & \textbf{1.24} \\ 
        \midrule
        & IM-VQG~\cite{krishna2019information} & 40.51 & 23.02 & 14.16 & 9.79 & 19.41 & 40.85 & 0.46 \\ 
        & TextRS-VQG~\cite{bashmal2023visual} & 37.38 & 26.09 & 19.71 & 15.52 & 17.38 & 37.93 & 0.69 \\ 
        TextRS-300 & LMQG~\cite{ushio2022generative} & 32.89 & 24.53 & 19.01 & 14.42 & \textbf{22.54} & 32.35 & \textbf{1.47} \\ 
        & ConVQG~\cite{mi2024convqg} & 43.06 & 33.38 & 27.61 & \textbf{23.20} & 19.67 & 42.60 & 1.35 \\ 
        & \textbf{KRSVQG} & \textbf{44.26} & \textbf{33.78} & \textbf{27.63} & 22.90 & 19.64 & \textbf{42.90} & \textbf{1.47} \\ 
        \bottomrule
    \end{tabular}}
\caption{\label{tab:finetune-rsdataset} Comparison of the question generation performance on the NWPU-300 and TextRS-300 datasets.}
\end{table*}

\begin{table*}[t!]
    \centering
    {\renewcommand{\arraystretch}{1.2}
    \begin{tabular}{c|c|ccc|ccccccc}
    \toprule
        Settings & Model & VPT & LPT & FT & BLEU 1 & BLEU 2 & BLEU 3 & BLEU 4 & METEOR & ROUGE$_{L}$ & CIDEr \\ 
        \midrule
        \multirow{5}{*}{\shortstack{Training\\ Strategy}} & KRSVQG w/o FT & \checkmark & \checkmark & $\times$ & 16.04 & 6.73 & 3.42 & 1.82 & 6.22 & 14.34 & 0.16 \\ 
        & KRSVQG w/o VPT & $\times$ & \checkmark & \checkmark & 34.89 & 22.99 & 16.83 & 11.99 & 15.61 & 35.80 & 0.92 \\ 
        & KRSVQG w/o LPT & \checkmark & $\times$ & \checkmark & 32.46 & 19.51 & 12.57 & 8.47 & 14.67 & 32.44 & 0.86 \\ 
        & \textbf{KRSVQG} & \checkmark & \checkmark & \checkmark & \textbf{41.87} & \textbf{28.32} & \textbf{20.73} & \textbf{14.78} & \textbf{18.70} & \textbf{38.48} & \textbf{1.24} \\ 
        \midrule
        \multirow{2}{*}{Structure} & KRSVQG w/o CGD & $\times$ & \checkmark & \checkmark & 34.77 & 21.49 & 14.62 & 9.98 & 15.32 & 35.00 & 0.92 \\
        & \textbf{KRSVQG} & \checkmark & \checkmark & \checkmark & \textbf{41.87} & \textbf{28.32} & \textbf{20.73} & \textbf{14.78} & \textbf{18.70} & \textbf{38.48} & \textbf{1.24} \\ 
        \bottomrule
    \end{tabular}}
\caption{\label{tab:ablation} Ablation study for training strategy and architecture components on the NWPU-300 dataset, where KRSVQG w/o VPT, KRSVQG w/o LPT, KRSVQG w/o FT, and KRSVQG w/o CGD represent the KRSVQG model without vision pre-training process, language pre-training process, fine-tuning on the RS dataset, and caption decoder ($\mathbf{BERT_{CapDec}}$), respectively. The best performance is in bold.}
\end{table*}

\subsection{Main Results}
In this part, we report and discuss the results achieved by the model at each stage of the training, accompanied by a comparison with state-of-the-art methods.

\subsubsection{VPT}
The evaluation results of VPT on the image caption generation task, using the NWPU caption dataset~\cite{cheng2022nwpu} are presented in Table~\ref{tab:nwpu-caption}. In this context, we focus on pre-training the image encoder and caption decoder, which align with the BLIP image captioning model. 
Notably, when compared against other state-of-the-art caption generation models focused on remote sensing, the vision module in KRSVQG exhibits a remarkable improvement (12.00\% on BLEU-1 and 0.81\% on CIDEr), surpassing other models in all assessment metrics. 
These results indicate the effectiveness of VPT in extracting useful information from the images.

\subsubsection{LPT}
Table~\ref{tab:finetune-kvqg} presents the question generation results on the K-VQG dataset. The proposed KRSVQG model achieves competitive results compared to state-of-the-art methods, which demonstrates the effectiveness of KRSVQG in knowledge-aware question generation.

\subsubsection{Knowledge-aware Question Generation}
Results on the two proposed datasets are reported in Table~\ref{tab:finetune-rsdataset}. From the results, we can make the following observations: 

\begin{itemize}[leftmargin=*]
\item \textbf{VQG Baselines}. In both datasets, IM-VQG~\cite{krishna2019information} exhibits inferior performance when compared to other methods. This may be because the original model does not take external knowledge as input, even though we included external knowledge sentences. The results highlight the importance of effectively encoding these knowledge sentences for the task. Similar observations can be done for the TextRS-VQG model~\cite{abdullah2020textrs}, which is designed for remote sensing visual question generation. However, since the knowledge input is not part of the model design, the performance might be influenced negatively.

\item \textbf{QG Baseline}. The LMQG~\cite{ushio2022generative} model outperforms the aforementioned two VQG baselines, but it is still inferior to our proposed KRSVQG. As a language model, LMQG does not use images as input and only uses captions and knowledge sentences. The results confirm the importance of integrating knowledge information, but also reveal the limitations due to the absence of visual information.

\item \textbf{Knowledge-aware VQG Baseline}. ConVQG~\cite{mi2024convqg} is proposed to generate knowledge-aware questions for natural images. Compared to ConVQG, KRSVQG takes image captioning as an intermediate task to ensure image grounding of the generated questions, which shows superiority in the remote sensing domain.

\item \textbf{Our KRSVQG model outperforms all the competing models on most metrics}, with an improvement of $7.74$\% improvement on BLEU-1 on the NWPU-300 dataset and $1.20$\% for the TextRS-300 dataset. \textcolor{Black}{The results demonstrate the effectiveness of the proposed method over remote sensing VQG and knowledge-aware VQG baselines, indicating that KRSVQG can effectively capture the image content and key concepts from the knowledge sentence.} 
The importance of every learning stage (VPT, LPT, and FT) and model components is assessed in Section~\ref{sec:ablation}.

\item Finally, we observe that the overall results on the TextRS-300 dataset are superior to those on the NWPU-300 dataset. The reason might be linked to the shorter captions in the TextRS-300 dataset, as indicated in Table~\ref{tab:RS-anno}. The brevity of these captions may cause the annotated questions to rely more on the knowledge content, making them easier to predict based on the knowledge sentence inputs.

\end{itemize}

\subsection{Ablation Study}
\label{sec:ablation}
To demonstrate the importance of each step in our training strategy and of the components of the model architecture, we performed a series of ablation studies. 
We also test the model performance in low-data regimes, with only 25\%, 50\%, and 75\% of training data used for training.

\subsubsection{Training Strategy Ablations}

The results of the training strategy ablation are presented in Table~\ref{tab:ablation}, demonstrating that omitting any of these training steps leads to a decline in performance.
KRSVQG without FT shows a significant reduction in performance across all metrics (BLUE-1 drops 25.83\%). This emphasizes that fine-tuning the model on the target dataset is pivotal. It allows the model to specialize in the task and to capture nuances of terminology specific to the remote sensing domain.
Comparing two pre-training stages, KRSVQG without LPT exhibits lower values across all evaluation metrics compared with the KRSVQG without VPT (BLEU-1 drops 2.43\%). The results indicate that LPT provides the ability to integrate external information into the generated questions, which is crucial for the task.

\subsubsection{Structure Ablations} 
We excluded the caption generation decoder from KRSVQG.
Notably, there is an impressive relative 48.10\% improvement in BLEU-4 and a relative 34.78\% boost in CIDEr scores when the CGD is integrated. Our findings highlight the effectiveness of caption generation as an intermediate stage of image feature representation to facilitate the information transfer from image embeddings to language embeddings.

\begin{figure}[t!]
    \centering
    \includegraphics[width=9cm]{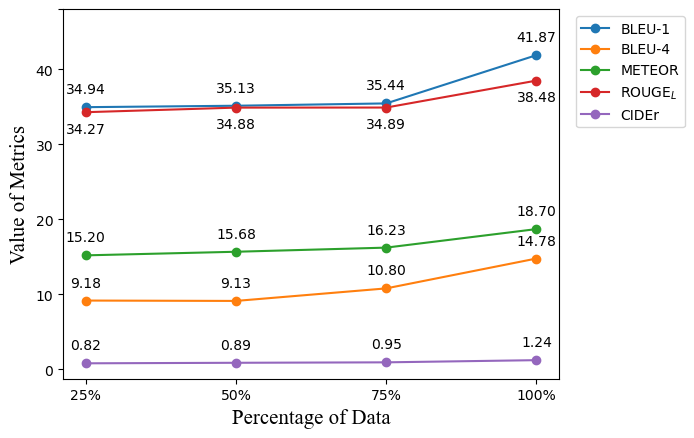}
    \caption{Performance comparison w.r.t. different amount of training data on the NWPU-300 dataset.}
    \label{fig:percent}
\end{figure}

\begin{figure*}[t!]
    \centering
    \includegraphics[width=18cm]{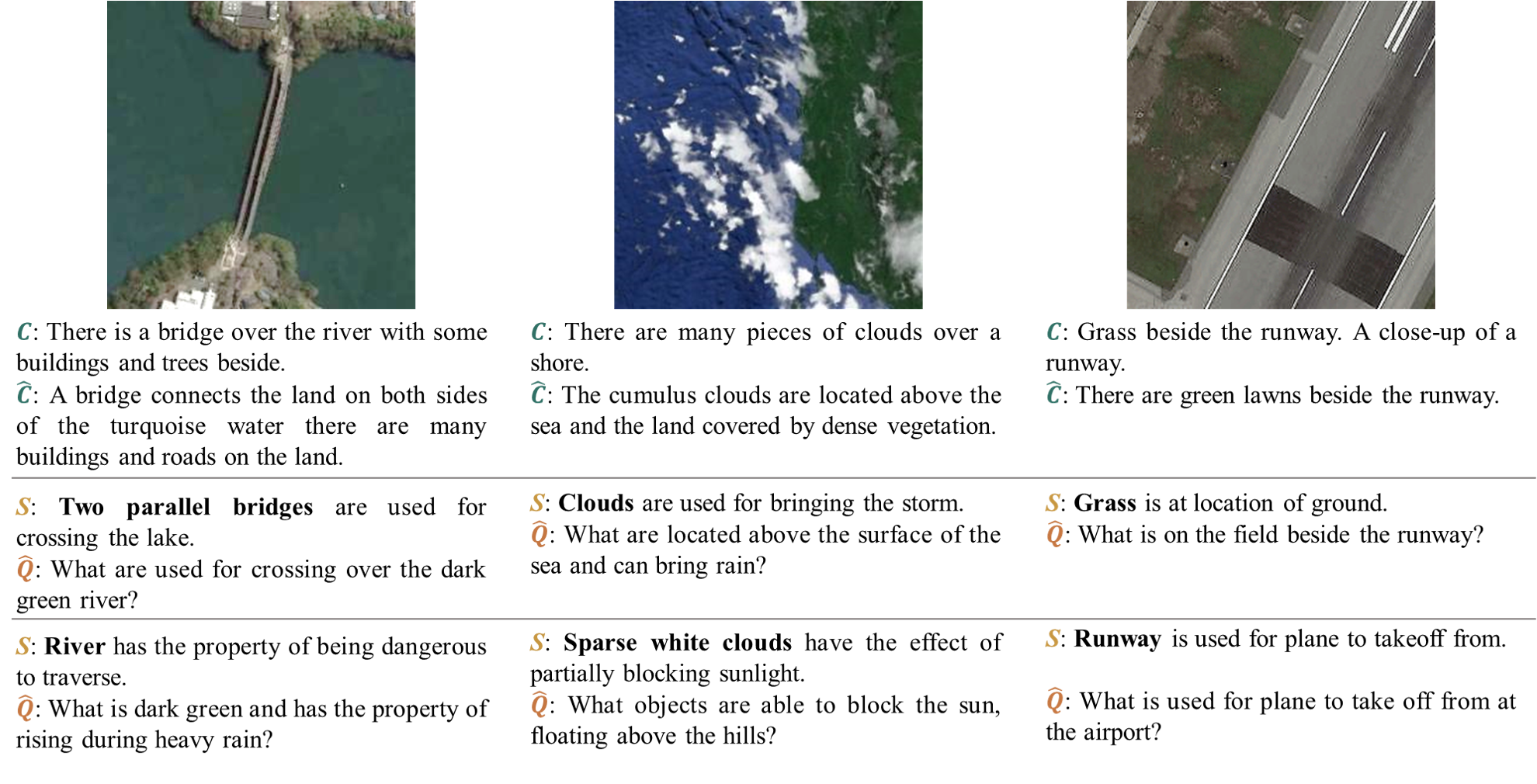}
    \caption{Questions generated by the KRSVQG model on the NWPU-300 dataset (left and middle) and TextRS-300 dataset (right). 
    Each sample contains the input image,  caption ($C$), generated caption ($\widehat{C}$), two knowledge sentences ($S$), and the corresponding generated questions ($\widehat{Q}$). 
    The corresponding answer for $\widehat{Q}$ is marked in bold.}
    \label{fig:nwpu_qg}
\end{figure*}

\subsubsection{Low-data Regime Analysis}
To assess our model's performance under data scarcity settings, we fine-tune the KRSVQG model using different subsets of the NWPU-300 training dataset, as shown in Fig.~\ref{fig:percent}. As expected, the model performs at its best when the largest amount of data is used for training. \textcolor{Black}{However, even when fine-tuned on a severely limited portion of the training data (25\%, only 60 images), our model's performance remains relatively consistent. The results demonstrate that a carefully annotated high quality dataset and a properly designed training strategy will contribute to the effectiveness of vision-language models, especially in remote sensing applications, where the low-resource setting is commonly encountered.}

\subsection{Qualitative Analysis}
In Fig.~\ref{fig:nwpu_qg}, we present examples of generated questions using the KRSVQG model. Our model excels at producing both descriptive captions $\widehat{C}$ and knowledge-aware questions $\widehat{Q}$, which integrate the information from both the visual content and the provided knowledge sentence $S$. Specifically, we provide two different knowledge sentences per image to the model. The results highlight the model's ability to generate a wide range of diverse questions based on varying knowledge inputs. For example, using the same image, different knowledge sentences such as (\textit{Clouds are used for bringing the storm.}) and (\textit{Sparse white clouds have the effect of partially blocking sunlight.}), allow us to create questions from different perspectives.

\begin{figure}[t!]
    \centering
    \includegraphics[width=9cm]{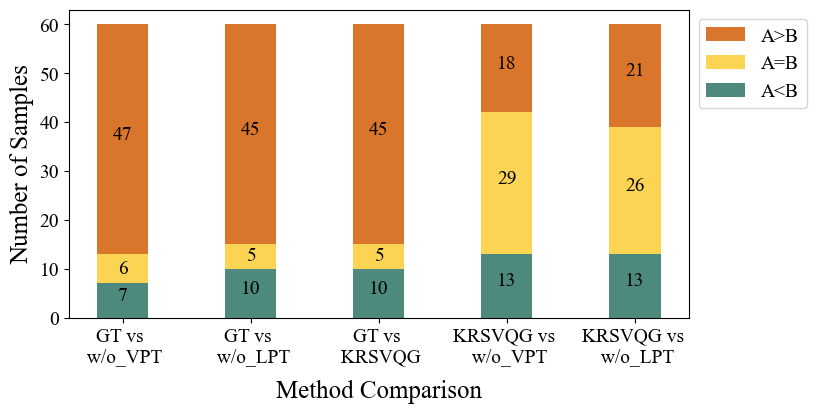}
    \caption{\textcolor{Black}{Results of human evaluation. GT, w/o\_VPT, w/o\_LPT, KRSVQG represent ground truths, questions from KRSVQG w/o VPT, KRSVQG w/o LPT, and KRSVQG model, respectively. $A>B$, $A=B$, and $A<B$ mean the generated question from the first model is better than, equal to, and worse than the ones generated from the second model, respectively.}}
    \label{fig:human-eval}
\end{figure}

\subsection{Human Evaluation}

Beyond the quantitative metrics evaluating the performance of the models, human evaluation is critical for assessing the contextual appropriateness of the generated questions. We conduct a human evaluation of the generated questions from different models on the NWPU-300 validation dataset with 60 images. For each image, there are four questions to be compared pair-wise: the questions generated by 1) our KRSVQG model, 2) KRSVQG without VPT, 3) KRSVQG without LPT, and 4) the ground truth questions. Ten English speakers are invited to participate in the evaluation. They are asked to compare a pair of generated questions for the same given image and to select the more appropriate question based on image relevance, answer, and external knowledge alignment.

In Fig.~\ref{fig:human-eval}, we present the results of human evaluation. While our models demonstrate strong performance, they still do not surpass the ground truth. However, it is noteworthy that 25\% of the questions generated by our KRSVQG model match or even outperform the ground truth. Moreover, a notable distinction emerges within KRSVQG variants. Approximately 78\% of the questions generated by the KRSVQG model are equal to or surpass those generated by the variants without VPT or LPT. These outcomes demonstrate the effectiveness of our training strategy in enhancing the quality of question generation.

\section{Conclusion \label{conclusion}}

In this paper, we propose a novel approach and benchmark datasets for generating knowledge-aware visual questions for remote sensing images. The KRSVQG model incorporates knowledge triplets into the question generation process and uses image captioning as an intermediate stage for image representation. Additionally, our proposed training strategy adapts the model to achieve the task with limited annotations, which provides a possible solution for low-data scenarios in real-world applications.
Results on two knowledge-aware remote sensing question generation datasets demonstrate the effectiveness of KRSVQG, which surpasses state-of-the-art VQG and QG methods in producing image-grounded and knowledge-enriched questions, instead of receiving the constraints of pre-defined templates. By moving beyond using only the information in pixels, knowledge-aware question generation could inspire and support future research efforts, potentially leading to more commonsense-enriched and contextually informed vision-language models.

\ifCLASSOPTIONcaptionsoff
  \newpage
\fi

\bibliographystyle{IEEEtran}
\bibliography{reference}

\end{document}